\documentclass[sigconf]{acmart}

\usepackage{booktabs}
\usepackage{subcaption}
\usepackage{graphicx}
\usepackage{amsmath}
\usepackage{caption}
\usepackage{array}
\usepackage{multirow}
\usepackage{xcolor}
\usepackage{makecell}
\usepackage{tabularx}
\newcolumntype{Y}{>{\centering\arraybackslash}X}
\usepackage{subfiles}
\usepackage{hyperref}

\AtBeginDocument{%
  }

\copyrightyear{2025}
\acmYear{2025}
\setcopyright{acmlicensed}\acmConference[MM '25]{Proceedings of the 33rd
ACM International Conference on Multimedia}{October 27--31, 2025}{Dublin,
Ireland}
\acmBooktitle{Proceedings of the 33rd ACM International Conference on
Multimedia (MM '25), October 27--31, 2025, Dublin, Ireland}
\acmDOI{10.1145/3746027.3755782}
\acmISBN{979-8-4007-2035-2/2025/10}

\begin{document}

\title{Drawing2CAD: Sequence-to-Sequence Learning for CAD Generation from Vector Drawings}

\author{Feiwei Qin}
\orcid{0000-0001-5036-9365}
\authornote{Equal contributions.}
\affiliation{%
  \institution{Hangzhou Dianzi University}
  \city{Hangzhou}
  \country{China}
}
\email{qinfeiwei@hdu.edu.cn}

\author{Shichao Lu}
\orcid{0009-0008-8017-1374}
\authornotemark[1]
\affiliation{%
  \institution{Hangzhou Dianzi University}
  \city{Hangzhou}
  \country{China}
}

\email{lushichao@hdu.edu.cn}

\author{Junhao Hou}
\orcid{0000-0003-1019-1595}
\authornotemark[1]
\affiliation{%
  \institution{Zhejiang University}
  \city{Hangzhou}
  \country{China}
}
\email{junhaohou@zju.edu.cn}

\author{Changmiao Wang}
\orcid{0000-0003-2466-5990}
\affiliation{%
 \institution{Shenzhen Research Institute of Big Data}
 \city{Shenzhen}
 \country{China}
}
\email{cmwangalbert@gmail.com}

\author{Meie Fang}
\orcid{0000-0003-4292-8889}
\authornote{Corresponding authors.}
\affiliation{%
  \institution{Guangzhou University}
  \city{Guangzhou}
  \country{China}
}
\email{fme@gzhu.edu.cn}

\author{Ligang Liu}
\orcid{0000-0003-4352-1431}
\authornotemark[2]
\affiliation{%
  \institution{University of Science and Technology of China}
  \city{Hefei}
  \country{China}
}
\email{lgliu@ustc.edu.cn}

\renewcommand{\shortauthors}{Feiwei Qin et al.}

\begin{abstract}

Computer-Aided Design (CAD) generative modeling is driving significant innovations across industrial applications. Recent works have shown remarkable progress in creating solid models from various inputs such as point clouds, meshes, and text descriptions. However, these methods fundamentally diverge from traditional industrial workflows that begin with 2D engineering drawings. The automatic generation of parametric CAD models from these 2D vector drawings remains underexplored despite being a critical step in engineering design. To address this gap, our key insight is to reframe CAD generation as a sequence-to-sequence learning problem where vector drawing primitives directly inform the generation of parametric CAD operations, preserving geometric precision and design intent throughout the transformation process. We propose Drawing2CAD, a framework with three key technical components: a network-friendly vector primitive representation that preserves precise geometric information, a dual-decoder transformer architecture that decouples command type and parameter generation while maintaining precise correspondence, and a soft target distribution loss function accommodating inherent flexibility in CAD parameters. To train and evaluate Drawing2CAD, we create CAD-VGDrawing, a dataset of paired engineering drawings and parametric CAD models, and conduct thorough experiments to demonstrate the effectiveness of our method. Code and dataset are available at \url{https://github.com/lllssc/Drawing2CAD}.

\end{abstract}


\begin{teaserfigure}
  \centering
  \includegraphics[width=0.80\textwidth]{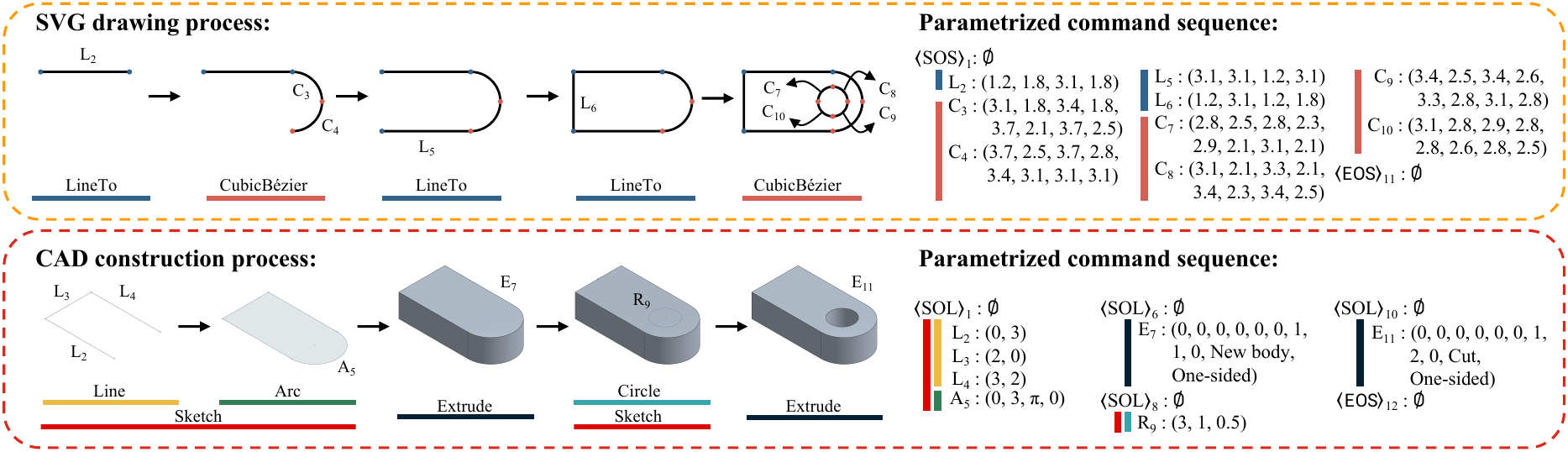}
  \caption{An intuitive comparison of SVG drawing and CAD construction processes. Both SVG drawing and CAD construction rely on a specific set of commands, and their respective processes can be formally represented as parametric command sequences with a unified structural format.}
  \label{fig:teaser}
\end{teaserfigure}


\maketitle

\section{Introduction}

In modern design and engineering, Computer-Aided Design (CAD) models are essential for product definition and iteration, playing a key role in prototyping, simulations, and manufacturing processes \cite{zou2024intelligent, zhou2025status, zhang2025ecad}. Parametric CAD modeling, which defines models as sequences of operations with geometric constraints, has become the industry standard due to its flexibility in enabling rapid design iterations through parameter adjustments \cite{willis2021fusion, wu2021deepcad, zhou2023cadparser, li2024cad}. The industrial design workflow typically initiates with engineering drawings, which function as the primary medium for conveying design intent and establishing the foundation for subsequent development stages \cite{tovey1989drawing, fan2025history, camba2022sketch, furferi20102d, niu2025pht}. Designers then create 3D parametric CAD models based on these 2D drawings to continue the product development process. This modeling process involves complex manual operations in CAD tools, requiring considerable time and expertise \cite{hu2023plankassembly, zhang2023automatic, wang20242d}, thus limiting design efficiency. This paper focuses on the specific task of generating parametric CAD models from vector engineering sketches represented in Scalable Vector Graphics (SVG) format.

Current research has explored generating parametric CAD models from point clouds \cite{uy2022point2cyl,liu2024point2cad}, meshes \cite{shen2025mesh2brep}, voxels \cite{lambourne2022reconstructing}, images \cite{zhang2024view2cad, chen2024img2cad}, and text descriptions \cite{li2024cad}. However, these approaches deviate from industrial design workflows that begin with 2D engineering drawings. While text-to-CAD generation has gained recent research attention, text descriptions struggle with precise dimensional specifications and spatial relationships. In contrast, sketches inherently excel at conveying geometric constraints, offering an intuitive medium for expressing design intent that directly addresses the ambiguity of text-based approaches.

Despite these advantages of sketches, existing sketch-based CAD generation approaches predominantly utilize raster image inputs \cite{li2022free2cad,harish2021photo2cad,chen2024img2cad}, facing fundamental limitations in extracting precise geometric information from pixels. These methods struggle with scale invariance, line thickness variations, and differentiating between design elements and rasterization artifacts, thus compromising geometric accuracy and design intent preservation. Vector formats such as SVG present an unexplored opportunity, as they inherently encode precise geometric primitives that align with engineering design intent. Nevertheless, generating parametric CAD models from SVG sketches introduces three key challenges: (1) preprocessing and parsing SVG files to extract meaningful geometric information \cite{han2020spare3d, yavartanoo2024text2cad}; (2) bridging the dimensional gap through cross-modal synthesis that transforms 2D vector sketches into 3D parametric CAD models; and (3) the absence of standardized SVG-to-CAD datasets.

To address these challenges, we introduce Drawing2CAD, a novel framework that enables cross-modal generation from 2D vector drawings to parametric CAD models.
The basic idea is illustrated in Figure~\ref{fig:teaser}. We redefine CAD model generation as a sequence-to-sequence learning problem, where our framework encodes vector drawing primitives from SVG sketches and synthesizes corresponding parametric CAD operations while preserving geometric precision and design intent. Specifically, Drawing2CAD operates in three key stages: (1) we first develop a network-friendly representation for vector engineering drawings that embeds SVG primitives while preserving precise geometric information and spatial relationships; (2) these embedded representations are then fed into our proposed dual-decoder architecture, where the first decoder generates CAD command types while the second decoder produces corresponding parameters, with command-guided generation, ensuring contextually appropriate parameters through this task-specialized decomposition; and (3) our end-to-end framework is optimized using a novel soft target distribution loss function that acknowledges inherent flexibility in CAD parameters, allowing subtle variations while preserving design intent.

Our approach offers several key advantages over existing methods. First, by using vector drawings as input, our method aligns naturally with standard CAD design workflows, where engineers typically begin with 2D sketches before proceeding to 3D modeling. Second, our direct processing of SVG primitives preserves the precise geometric information and relationships critical for accurate CAD modeling, enabling high-fidelity representation of design intent compared to raster image-based approaches. Third, the framework accepts flexible input configurations (a single isometric view, three orthographic views, or all four views combined), making it widely applicable across different application scenarios and adaptable to diverse designer preferences. These advantages collectively enable our approach to generate high-quality parametric CAD models that maintain engineering fidelity while supporting efficient design workflows.

To support this research, we create CAD-VGDrawing, a large-scale dataset containing over 150,000 CAD models and their corresponding engineering drawings in both vector and raster formats. Experimental results demonstrate that vector engineering drawings provide a more suitable and information-rich input for CAD operation sequence generation compared to raster-based representations, underscoring the advantages of vector graphics in advancing 3D CAD modeling. Furthermore, our method surpasses baseline approaches, highlighting its effectiveness in generating CAD operation sequences that better align with the original design intent, thereby improving overall model performance.

In summary, this work has the following contributions:
\begin{itemize}
    \item We develop a network-friendly representation for vector engineering drawings that preserves precise geometric information and spatial relationships, enabling deep learning models to effectively process structured SVG primitives.
    \item We introduce Drawing2CAD, a novel framework for cross-modal generation from 2D vector engineering drawings to parametric CAD models, redefining sketch-based CAD generation as a sequence-to-sequence learning problem that directly processes SVG primitives to preserve precise geometric information and spatial relationships.
    \item We propose a novel dual-decoder architecture that decomposes the complex CAD generation task into command type prediction and parameter estimation, incorporating a soft target distribution loss function to enable more precise geometric control and enhance the effectiveness of the generation model.
    \item We create CAD-VGDrawing, a comprehensive dataset containing paired vector engineering drawings and parametric CAD models, and demonstrate significant performance improvements of our approach compared to baseline methods in command accuracy, parameter precision, and CAD model validity.
\end{itemize}

\section{Related Work}

\noindent
\subsection{Parametric CAD Modeling}
Parametric CAD modeling has evolved significantly with deep learning approaches for generating operation sequences, progressing from foundational works \cite{willis2021fusion, wu2021deepcad} to advanced techniques leveraging Transformer architectures \cite{xu2022skexgen} and diffusion models \cite{zhang2025diffusion} that enhance geometric consistency. Current research primarily explores reconstruction from existing 3D representations: point clouds \cite{ma2023multicad, dupont2024transcad, khan2024cad, ma2024draw, uy2022point2cyl, liu2024point2cad}, boundary representations \cite{zhou2023cadparser, zhang2024brep2seq, zhang2025ecad}, and multi-view images \cite{zhang2024view2cad, alam2024gencad, li2025image2cadseq, chen2024img2cad}, each employing modality-specific encoders \cite{qi2017pointnet++,lambourne2021brepnet,dosovitskiy2020image} to map inputs into latent vectors for sequence decoding. However, these approaches mainly focus on reverse engineering scenarios where 3D objects already exist. Text-to-CAD methods \cite{li2024cad, yavartanoo2024text2cad} attempt forward design but lack the precision of engineering drawings. Despite vector engineering drawings serving as the starting point for industrial design workflows, no prior research has explored generating parametric CAD sequences directly from these representations.

\noindent
\subsection{Sketch-Based CAD Modeling}
Sketch-based CAD modeling has progressed through several parad-\\igms, from early mathematical frameworks and geometric reasoning approaches \cite{wang1993survey, kuo1998reconstruction, liu2001reconstruction, gong2006reconstruction, furferi20102d} that established theoretical foundations but were limited to simple geometries, to feature recognition methods \cite{harish2021photo2cad, camba2022sketch} for extracting geometric patterns from engineering drawings, and finally to contemporary image-based deep learning approaches that explored bitmap-based sketches through convolutional networks \cite{puhachov2023reconstruction, wang20242d}, stroke decomposition \cite{li2022free2cad}, and pattern matching \cite{zhang2023automatic}. 
However, raster-based methods are fundamentally constrained when capturing engineering precision, encountering difficulties with resolution dependence, feature extraction accuracy, and geometric fidelity preservation. The potential of vector graphic formats remains largely unexplored in CAD generation research, despite their natural representation of geometric elements and widespread adoption in engineering design tools, highlighting an opportunity for our approach that leverages the structural and geometric advantages of SVG for direct CAD operation sequence generation.

\noindent
\subsection{Vector Graphics in Deep Learning}
In the field of computer graphics, two predominant image formats prevail: raster images, characterized by pixel matrices, and vector images, such as Scalable Vector Graphics (SVG), characterized by a series of code language commands \cite{tang2024strokenuwa}.
A pioneering work in vector graphics generation, DeepSVG \cite{carlier2020deepsvg}, introduced a hierarchical Transformer-based generative model specifically for vector graphics generation. It also curated a large-scale SVG dataset and integrated deep learning techniques for SVG manipulation and editing, laying the groundwork for our study. 
Subsequent research has made great achievements in SVG representation learning \cite{reddy2021multi,campbell2014learning,lopes2019learned} and generative models \cite{reddy2021im2vec,wang2021deepvecfont}.
However, existing studies have primarily focused on traditional vector graphic applications, such as fonts, icons, and digital illustrations, without exploring the intersection of vector graphics and CAD engineering drawings. To explore the potential synergies arising from the intersection of these two fields, we leverage vector graphics techniques in deep learning and, for the first time, propose a method to generate CAD operation sequences from vector engineering drawings.

\section{Preliminary}

For easy to understand the rational design of Drawing2CAD, we first introduce the concepts of CAD operation sequences and SVG drawing sequences. CAD operation sequences are textual representations comprising commands and parameters essential for computer-aided design. They enable complex object construction through parameter manipulation and primitive shape combination (cubes, spheres, cylinders). This parametric approach allows automatic model updates when parameters change. Our research focuses on generating single objects using specific commands: "Line", "Circle", "Arc", and "Extrude". Similarly, SVG drawing sequences are textual representations used in scalable vector graphics, simplifying graphical modeling into discrete commands—primarily "LineTo" for straight lines and "Cubic Bézier" for smooth contours. Like CAD sequences, SVG sequences enable design modifications through parameter adjustments, providing an intuitive mechanism for programmatic graphic generation. For detailed definitions and structures of both sequence types, see \cite{wu2021deepcad} and \cite{carlier2020deepsvg}.

\section{Method}

\begin{figure*}[t]
    \centering
    \includegraphics[width=\textwidth]{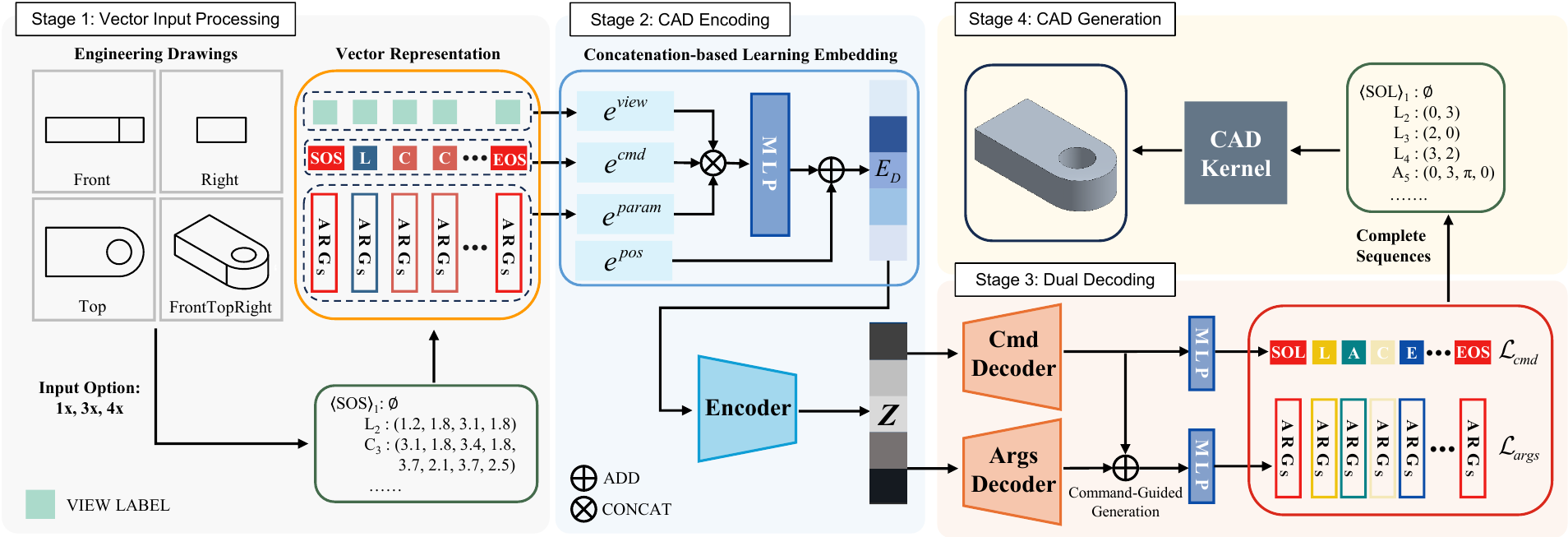}
    \caption{The pipeline of our proposed method. Drawing2CAD takes vector engineering drawings in one of three view configurations as input, encodes them into a latent vector, and employs a dual-decoder to generate CAD command types and their parameters. The resulting complete operation sequences are processed by a CAD kernel to build the final 3D model.}
    \label{fig:pipeline}
\end{figure*}

\subsection{Overview}
We propose Drawing2CAD, a Transformer-based \cite{vaswani2017attention} network that takes vector engineering drawings as input in one of three optional forms (a single isometric view, three orthographic views, or a combination of all four views) and generates CAD operation sequences that meet design intent. The overall architecture is shown in Figure~\ref{fig:pipeline}.
We design a network-friendly representation for each vector engineering drawing. During the training stage, we first adopt embedding for vector engineering drawings before feeding them into the encoder. The latent vector outputted by the encoder is then passed through our designed dual-decoder architecture. The two decoders take the latent vector as input, generating the command and parameter parts of the CAD operation sequence, respectively. Based on the principle of one-to-one correspondence between command type and parameters, we use the generated commands to guide the generation of parameters. Finally, by merging the commands and parameters, we obtain the complete CAD operation sequences, which can be processed through the OpenCASCADE-based CAD kernel \cite{pythonocc} to generate the CAD model. Meanwhile, we improve the objective function from previous research, providing more suitable training constraints.

\subsection{CAD-VGDrawing Dataset}
While existing datasets provide CAD models with construction process records \cite{willis2021fusion, zhou2023cadparser} and vector representations of operation sequences \cite{wu2021deepcad}, they lack corresponding engineering drawings. To address this gap, we introduce CAD-VGDrawing, an SVG-to-CAD dataset that pairs engineering drawings with their corresponding CAD models and operation sequences.

\noindent
\textbf{Collection of Vector Engineering Drawings.}
The CAD models used in our study were sourced from the DeepCAD dataset \cite{wu2021deepcad}. To streamline the engineering drawing generation process, we developed a custom Python script in FreeCAD \cite{freecad}. This script utilizes FreeCAD's TechDraw module to automatically convert imported STEP files into engineering drawings and export them as SVG files. For each model, we generated four standard views: three orthographic projections (front, top, and right) and one isometric (front-top-right) view. We then used CairoSVG \cite{cairosvg} to convert these vector drawings to PNG format. This rasterization step enables integration with image-based learning pipelines and supporting comparative analysis between vector and raster representations as input sources.

\noindent
\textbf{Vector Engineering Drawings Preprocessing.}
We performed systematic preprocessing on all engineering drawings. First, we attempted to simplify Bézier curve segments following Carlier et al. \cite{carlier2020deepsvg}, but found FreeCAD's native curve segmentation already optimal through the comparative analysis, requiring no further refinement. Next, we addressed path ordering issues by implementing path reordering, as FreeCAD typically generates SVG paths in an irregular and inconsistent sequence. To standardize path ordering, we used the canvas top-left corner as origin and applied graph theory algorithms \cite{networkx} to identify all contours. These contours were then arranged by increasing distance from origin, with each contour drawn clockwise. Finally, we normalized all SVG drawings to a standardized $200 \times 200$ viewbox. For the bitmap-format engineering drawings, we uniformly resized all PNG images to $224 \times 224$, ensuring consistency for image-based learning tasks and facilitating standardized input processing. These preprocessing steps enabled accurate geometric information extraction for our representation method, detailed in Section~\ref{sec:representation}.

\noindent
\textbf{Statistics of CAD-VGDrawing Dataset.}
Our automated preprocessing workflow handled approximately 176,000 CAD models from the DeepCAD dataset. However, FreeCAD encountered limitations when processing a subset of complex models, producing either invalid engineering drawings or representations that significantly deviated from the original CAD models. To ensure dataset quality and reliability, we systematically filtered these problematic instances, resulting in a curated dataset of 161,407 CAD models with corresponding engineering drawings across four distinct views.
A detailed statistical analysis was performed on our dataset, covering three aspects: (1) the distribution of engineering drawings categorized by SVG command types (e.g., LineTo, CubicBézier, or their combination), (2) the sequence length distribution of SVG drawing commands, and (3) the CAD sequence length distributions in the retained versus filtered subsets of the dataset. These results are presented in the supplementary material.
Based on this analysis, we selected SVG engineering drawings with sequence lengths not exceeding 100 to construct our final dataset. This strategic decision ensured data completeness and diversity while reducing the negative impact of variable-length sequences on model performance, thereby decreasing model complexity and improving stability. The final dataset comprised 157,591 SVG-to-CAD pairs, randomly divided into training (90\%), validation (5\%), and test (5\%) sets.

\subsection{Vector Engineering Drawings Representation}
\label{sec:representation}
Unlike previous approaches \cite{carlier2020deepsvg}, we adopt a simplified representation for vector engineering drawings by focusing exclusively on core geometric information. We exclude non-essential path attributes (visibility, color, fill properties) and restrict our command set to LineTo (L) and CubicBézier (C). This approach standardizes each command to an 8-value parameter list:
\begin{equation}
X=(x_1,y_1,c_{x1},c_{y1},c_{x2},c_{y2},x_2,y_2) \in \mathbb{R}^8,
\end{equation}
where $x_1,y_1$ and $x_2,y_2$ represent start and end points, while $c_{x1},c_{y1}$ and $c_{x2},c_{y2}$ are control points for Bézier curves. For LineTo commands, only the start and end points are utilized, with control point parameters set to -1, while CubicBézier commands employ all eight parameters.

Since each CAD model corresponds to four engineering drawings (Front, Top, Right, and Isometric views), we assign specific view labels to maintain proper contextual identification. Overall, we define each vector engineering drawing as an ordered sequence $D_i=\{S_1,S_2,...,S_N\}$, where $D_i$ represents the $i$-th engineering drawing containing $N = 100$ command sequences. Each sub-sequence $S_i=(v_i,C_i)$ includes a view label $v_i \in \{Front, Top, Right, Isometric\}$ and a command sub-sequence $C_i=(c_i^j,X_i^j)$. Here, $c_i^j$ is one of the elements in the command set $\{\langle SOS\rangle,L,C,\langle EOS \rangle \}$, which denotes the command type ($L$ for LineTo, $C$ for CubicBézier, with start and end sequence markers), and $X_i^j$ contains the 8-dimensional command parameters:
\begin{equation}
X_i^j=(q^j_{x_1,i},q^j_{y_1,i},q^j_{c_{x1},i},q^j_{c_{y1},i},q^j_{c_{x2},i},q^j_{c_{y2},i},q^j_{x_2,i},q^j_{y_2,i}).
\end{equation}
This representation enables us to encode the complete geometric information of engineering drawings while maintaining a consistent, standardized format across different command types.

\subsection{Architecture}
\textbf{Embedding.}
To effectively structure input for the Transformer network, we project the view labels, SVG commands, and parameters into a continuous embedding space of dimension $d_E$. Unlike prior methods \cite{carlier2020deepsvg, wu2021deepcad, zhou2023cadparser, li2024cad} that fuse embeddings via direct linear addition, treating command and parameter information independently, our approach explicitly models the interactions among the view, command, and parameter components through concatenation-based embedding learning. This design choice is supported by our Ablation Study~\ref{ablation study}. Specifically, we concatenate the respective embeddings and apply a linear transformation (MLP) to produce a unified fused representation, which enables richer cross-field interactions:
\begin{equation}
E_D(i) = \boldsymbol{MLP}(\boldsymbol{CONCAT}(e_i^{view}, e_i^{cmd}, e_i^{param})) + e_i^{pos} \in \mathbb{R}^{d_E}.
\end{equation}
\noindent
The view embedding $e_i^{view}$ encodes the view type $v_i$ as: $e_i^{view} = W_{view} \delta_i^v$,
where $W_{view} \in \mathbb{R}^{d_e \times 4}$ is a learnable matrix, and $\delta_i^v \in \mathbb{R}^4$ is a one-hot vector indicating one of four standard views.
The command embedding $e_i^{cmd}$ represents the command type $c_i$ as:
$e_i^{cmd} = W_{cmd} \delta_i^c$,
where $W_{cmd} \in \mathbb{R}^{d_e \times 4}$ is a learnable matrix, and $\delta_i^c \in \mathbb{R}^4$ is a one-hot encoding of the predefined command set.
The parameter embedding $e_i^{param}$ encodes the command parameters. As described in Section~\ref{sec:representation}, each command contains eight parameters quantized into 8-bit integers. Each integer is converted into a one-hot vector $\delta_{i,j}^p$ ($j = 1, \dots, 8$) of dimension $2^8 + 1 = 257$, with the extra dimension accommodating unused parameters. These vectors form a matrix $\delta_i^p \in \mathbb{R}^{257 \times 8}$.
To embed each parameter, we apply a shared learnable matrix $W_{param}^b \in \mathbb{R}^{d_e \times 257}$ column-wise, and then flatten the resulting embeddings before passing them through a linear projection $W_{param}^a \in \mathbb{R}^{d_e \times 8d_e}$:
\begin{equation}
    e_i^{param} = W_{param}^a \, \text{flat}(W_{param}^b \delta_i^p),
\end{equation}
where $\text{flat}(\cdot)$ flattens the input matrix into a vector. The $\boldsymbol{MLP}(\cdot)$ learns cross-field interactions through a linear projection. When only the isometric view serves as input, the view embedding becomes redundant due to the lack of variation, and the overall embedding will be simplified:
\begin{equation}
E_D(i) = \boldsymbol{MLP}(\boldsymbol{CONCAT}(e_i^{cmd}, e_i^{param})) + e_i^{pos} \in \mathbb{R}^{d_E}.
\end{equation}
\noindent
The function of positional encoding $e_i^{pos}$ is the same as in the original Transformer \cite{vaswani2017attention}, which is used to record the index of the command in the complete vector engineering drawing SVG sequence. In our implementation, the dimension of $d_E$ is set to 256. For the multi-view setting, we adopt a straightforward strategy by stacking the three orthographic (front, top, and right) views followed by the isometric (front-top-right) view in a fixed order.

\noindent
\textbf{Encoder.}
Our encoder $E$ consists of four Transformer blocks, each containing eight attention heads and a feed-forward dimension of 512. The encoder $E$ takes the embedded sequence $[e_1, ..., e_{N_c}]$ as input and outputs a sequence of vectors $[\hat{e_1}, ..., \hat{e}_{N_c}]$, where each vector has the same dimension $d_E = 256$. Finally, the output vectors are averaged to produce a single $d_E$-dimensional latent vector $z$.

\noindent
\textbf{Dual Decoder.}
Our decoder adopts a dual-decoder architecture consisting of two independent Transformer decoders with identical hyper-parameter settings as the encoder. Both decoders take a learned constant embedding as input while attending to the latent vector $z$ to capture global features.
In CAD operations, each command type often requires a specific set of parameters, and even the same command may require different parameters depending on the context. To address this complexity, we enforce a one-to-one correspondence between command types and their parameters, and decompose the generation task accordingly: the Command Decoder predicts the CAD operation type $\hat{t}_i$ while the Argument Decoder generates the associated parameter vector $\hat{p}_i=[x, y, \alpha, f, r, \theta, \gamma, p_x, p_y, p_s, s, e_1, e_2, b, \mu]$. A key innovation in our architecture is the command-guided parameter generation. To ensure that the Argument Decoder generates parameters consistent with the command semantics, we add the output of the Command Decoder to the output of the Argument Decoder. This fusion injects command-level information into the parameter generation process, effectively enhancing the decoder’s capacity to produce contextually appropriate and semantically aligned parameters. The outputs from each decoder’s Transformer block are then projected through separate linear layers to obtain the predicted command and parameters respectively.
Finally, the operation command types and parameters are combined to form the complete CAD operation sequences. The generated CAD operation sequences are subsequently processed by an OpenCASCADE-based CAD kernel to build the final CAD model.

\subsection{Loss Function}
Our approach employs a composite loss function consisting of Command Loss $\mathcal{L}_{cmd}$ and Parameter Loss $\mathcal{L}_{args}$, similar to existing CAD operation sequence generation methods. However, we introduce a significant enhancement to the Parameter Loss component. Traditional approaches typically rely on hard classification, requiring exact matches between predictions and ground truth values. In contrast, our method recognizes that CAD operation parameters naturally tolerate minor variations that can maintain design intent while introducing beneficial diversity to the resulting models. Based on this insight, we formulated our Parameter Loss using soft target distributions rather than rigid one-hot encodings:
\begin{equation}
\mathcal{L}_{args} = - \sum^{N_c}_{i=1}\sum_{j=1}^{N_p}\sum_{k=1}^{C} \tilde{y}_k \log \left( \hat{y}_k \right),
\end{equation}
where $N_c$ denotes the command sequence length, $N_p$ represents the number of command parameters, and $C$ is the number of discrete categories for each parameter.
The term $\tilde{y}_k$ is a smoothed probability distribution, which assigns penalization weights for predictions deviating from the true parameter category:
\begin{equation}
\tilde{y}_k = \frac{e^{-\alpha |k - y|}}{Z},
\end{equation}
where $\alpha$ controls the strength of tolerance decay, $|k - y|$ is the distance between the predicted parameter category and the true parameter category, and $Z$ is a normalization factor ensuring that the probabilities sum to 1.
Smoothing weights are applied exclusively to categories within the range $[y - tolerance, y + tolerance]$,  with zero weights assigned to all other categories. In practice, we set $\alpha$ to 2.0 and the $tolerance$ to 3. This innovative loss formulation alleviates excessive penalties for predictions that slightly deviate from the ground truth but still fall within an acceptable range. By relaxing overly strict constraints, it enhances the model’s generalization ability. As demonstrated in our Ablation Study~\ref{ablation study}, this refinement leads to consistent improvements in parameter-related metrics.

\subsection{Metrics}
\noindent
\textbf{Command Accuracy.}
To evaluate the prediction accuracy of commands, we employ two metrics: Command Type Accuracy ($ACC_{cmd}$) and Parameter Accuracy ($ACC_{param}$).
The Command Type Accuracy ($ACC_{cmd}$) measures the correctness of predicted CAD command types:
\begin{equation}
ACC_{cmd}=\frac{1}{N_c}\sum_{i=1}^{N_c}\mathbb{I}[t_i=\hat{t}_i],
\end{equation}
where $N_c$ is the total length of the CAD command sequence, $t_i$ and $\hat{t}_i$ represent the ground truth and predicted command type for the $i$-th command, respectively. The indicator function $\mathbb{I}$ returns 1 when the condition is satisfied and 0 otherwise.
For commands with correctly predicted types, we further evaluate the correctness of command parameters, defined as:
\begin{equation}
ACC_{param}=\frac{1}{K}\sum_{i=1}^{N_c}\sum_{j=1}^{|\hat{p}_i|}\mathbb{I}[|p_{i,j}-\hat{p}_{i,j}|<\eta]\mathbb{I}[t_i=\hat{t}_i],
\end{equation}
where $K$ represents the total parameters count in correctly predicted commands, $p_{i,j}$ and $\hat{p}_{i,j}$ denote the ground truth and predicted values for the $j$-th parameter of the $i$-th command, respectively.
$\eta$ is a tolerance threshold, defining the acceptable error margin between predicted and ground truth parameters. The indicator function $\mathbb{I}[t_i=\hat{t}_i]$ ensures that parameter accuracy is only evaluated for correctly predicted command types. We set $\eta = 3$ to account for parameter quantization, consistent with the tolerance threshold in our loss function.

\noindent
\textbf{Shape Construction and Evaluation.}
When a CAD model is constructed from generated CAD operation sequences, we can convert it into point clouds by randomly sampling K points on its surface. In practice, we set $K=2000$. To measure the differences between a real shape and the predicted shape, we calculate the Mean Chamfer Distance (MCD) of them. Additionally, we report the Invalidity Ratio (IR), which quantifies the percentage of generated CAD operation sequences that fail to produce valid 3D shapes.

\section{Experiments}

\subsection{Experimental Setup}
To the best of our knowledge, we propose the first framework that generates parametric CAD models directly from vector engineering drawings. Given the absence of existing models or benchmark datasets for this task, there are currently no available methods for direct comparison. To comprehensively evaluate our approach, we design a series of experiments covering different aspects of the task.
We begin by comparing the effect of input formats, analyzing the performance differences between using vector (\textit{.svg}) engineering drawings and raster (\textit{.png}) engineering drawings as input. Following this, we assess the effectiveness of our method by comparing it against a modified baseline model under various evaluation metrics, and further perform a visual comparative analysis with traditional methods to better understand the differences in output quality and structural correctness. To further understand the internal mechanisms and the contribution of each component in our framework, we conduct a set of ablation studies. Moreover, we analyze several imperfect or failure cases to uncover current limitations of our method and provide directions for future research.

The experiments were conducted on one NVIDIA RTX 4090 GPU with a batch size of 256 under 200 epochs. The training process utilized the Adam optimizer with a learning rate of 0.001, incorporating linear warm-up for the first 2000 steps. We applied a dropout rate of 0.1 to all Transformer blocks and used gradient clipping with a threshold of 1.0 during backpropagation.

\begin{table}[ht]
    \centering
    \caption{The comparison of results with raster (DeepCAD-raster) and vector (DeepCAD-vector) engineering drawing inputs. ACC\textsubscript{cmd}, ACC\textsubscript{param} and IR are multiplied by 100\%. MCD is multiplied by $10^2$. $\uparrow$ means a higher metric value indicates better results. $\downarrow$ means a lower
    metric value indicates better results.}
    \label{tab:vector_vs._raster}
    \setlength{\tabcolsep}{4pt}
    \resizebox{\linewidth}{!}{
    \begin{tabular}{llcccc}
        \toprule
        \textbf{Input} & \textbf{Method} & \textbf{ACC\textsubscript{cmd} $\uparrow$} & \textbf{ACC\textsubscript{param} $\uparrow$} & \textbf{IR $\downarrow$} & \textbf{MCD $\downarrow$} \\
        \midrule
        \multirow{2}{*}{\parbox[t]{2.8cm}{isometric (1x)}}
            & DeepCAD-raster & 75.60 & 69.44 & 30.54 & 17.67 \\
            & DeepCAD-vector & \textbf{80.78} & \textbf{73.73} & \textbf{23.29} & \textbf{11.52} \\
        \midrule
        \multirow{2}{*}{\parbox[t]{2.8cm}{orthographic (3x)}}
            & DeepCAD-raster & 76.81 & 70.74 & 30.38 & 18.73 \\
            & DeepCAD-vector & \textbf{81.39} & \textbf{74.76} & \textbf{22.97} & \textbf{12.15} \\
        \midrule
        \multirow{2}{*}{\parbox[t]{2.8cm}{isometric +\\orthographic (4x)}}
            & DeepCAD-raster & 77.69 & 70.49 & 29.79 & 18.16 \\
            & DeepCAD-vector & \textbf{81.51} & \textbf{75.14} & \textbf{23.40} & \textbf{11.37} \\
        \bottomrule
    \end{tabular}
    }
\end{table}

\begin{figure}[!htbp]
  \centering
  \includegraphics[width=\linewidth]{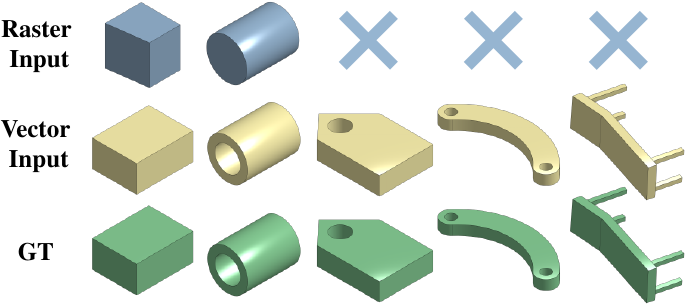}
  \caption{Comparison results with raster (DeepCAD-raster) and vector (DeepCAD-vector) engineering drawing inputs. "$\times$" means the generated parametric sequences that fail to reconstruct 3D shape.}
  \label{fig:vector_vs._raster}
\end{figure}

\subsection{Experimental Results}
\noindent
\textbf{Vector and Raster Inputs.}
To conduct our designed comparative experiments, we built a modified baseline model based on DeepCAD \cite{wu2021deepcad}.
We replaced the original CAD command input with SVG command input while keeping all other settings unchanged (referred to as "DeepCAD-vector").
For raster (\textit{.png}) engineering drawings, we replaced the original encoder with a SOTA Vision Transformer (ViT) pre-trained feature extractor \cite{oquab2023dinov2} (referred to as "DeepCAD-raster"). The detailed results of comparison between DeepCAD-vector and DeepCAD-raster are presented in Table~\ref{tab:vector_vs._raster}, which show that DeepCAD-vector outperforms DeepCAD-raster consistently, regardless of whether we use a single isometric view, three orthographic views, or all four views combined as input. Meanwhile, as illustrated in Figure~\ref{fig:vector_vs._raster}, when vector inputs are used, the model is able to generate CAD models that better align with the design intent. In contrast, with raster inputs, the generated CAD models often fail to meet design expectations and are more likely to result in invalid shapes.
These findings highlight the advantage of using vector inputs, which provide more precise and semantically rich structural information compared to raster pixel data. This allows the model to more effectively incorporate sketch design into its feature representation, thereby conveying more specific and meaningful design intent.

\noindent
\textbf{Performance Comparison.}
In our quantitative evaluation, we compared our approach with the baseline method DeepCAD-vector. As shown in Table~\ref{tab:performance}, our proposed method, Drawing2CAD, consistently outperforms DeepCAD-vector in all metrics, regardless of whether a single isometric view, three orthographic views, or all four views combined are used as input. When using only the isometric view, our method exhibits a slightly higher Mean Chamfer Distance (CD) compared to DeepCAD-vector. However, this difference should be considered alongside our significantly lower Invalidity Ratio (IR), as CD is only calculated for successfully generated models.

\begin{figure}[!htbp]
    \centering
    \includegraphics[width=\linewidth]{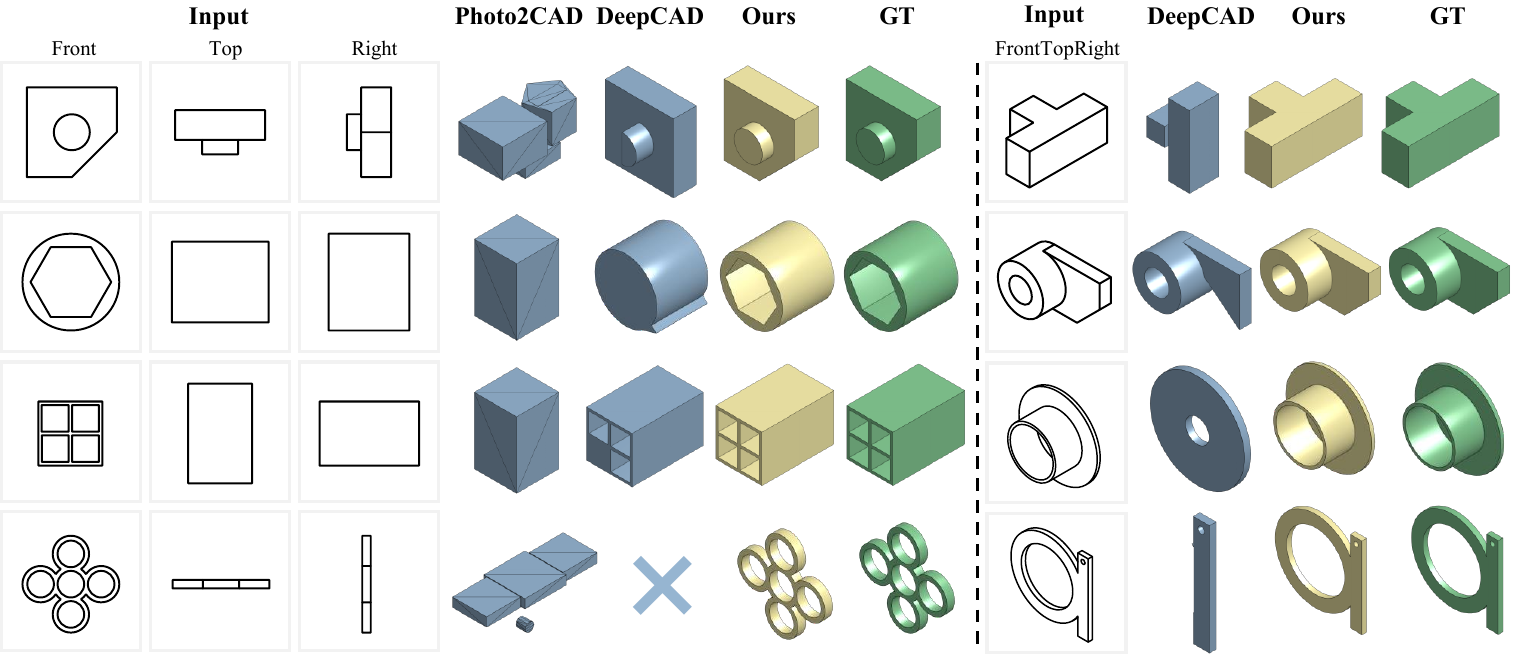}
    \caption{Comparison results of engineering drawings to parametric CAD models when using three orthographic views or a single isometric view as input.}
    \label{fig:1x and 3x}
\end{figure}

\begin{figure}[!htbp]
  \centering
  \includegraphics[width=\linewidth]{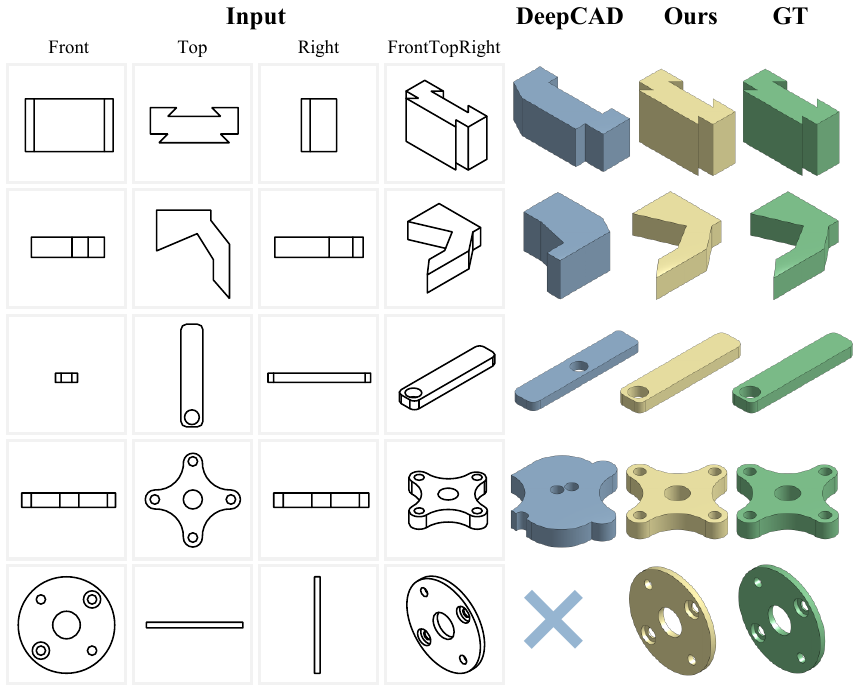}
  \caption{Comparison results of engineering drawings to parametric CAD models when using all four views as input.}
  \label{fig:4x}
\end{figure}

In our qualitative analysis, as illustrated in Figures~\ref{fig:1x and 3x} and~\ref{fig:4x}, Drawing2CAD generates CAD models that better align with the design intent conveyed in the engineering drawings, demonstrating clear improvements over the baseline methods. We also compared our approach with Photo2CAD \cite{harish2021photo2cad}, a traditional rule-based method that generates CAD models from orthographic drawings by extracting geometric features from three orthographic views, establishing hierarchical structures, and applying Boolean operations to create 3D models. As shown in Figure~\ref{fig:1x and 3x}, Photo2CAD struggles to effectively generate accurate CAD models for our test cases, particularly for complex geometries where rule-based approaches fail to capture the intricate design details. In contrast, our learning-based method successfully reconstructs these challenging models while preserving the geometric features and design intent specified in the input drawings.

\begin{table}[ht]
    \centering
    \caption{The comparison of DeepCAD-vector and Drawing2CAD on the vector engineering drawings to parametric CAD sequences generation.}
    \label{tab:performance}
    \setlength{\tabcolsep}{4pt}
    \resizebox{\linewidth}{!}{
    \begin{tabular}{llcccc}
        \toprule
        \textbf{Input} & \textbf{Method} & \textbf{ACC\textsubscript{cmd} $\uparrow$} & \textbf{ACC\textsubscript{param} $\uparrow$} & \textbf{IR $\downarrow$} & \textbf{MCD $\downarrow$} \\
        \midrule
        \multirow{2}{*}{\parbox[t]{2.8cm}{isometric (1x)}}
            & DeepCAD-vector & 80.78 & 73.73 & 23.29 & \textbf{11.52} \\
            & Drawing2CAD & \textbf{81.81} & \textbf{74.35} & \textbf{21.40} & \ 12.10 \\
        \midrule
        \multirow{2}{*}{\parbox[t]{2.8cm}{orthographic (3x)}}
            & DeepCAD-vector & 81.39 & 74.76 & 22.97 & 12.15 \\
            & Drawing2CAD & \textbf{82.12} & \textbf{75.43} & \textbf{20.99} & \textbf{11.98} \\
        \midrule
        \multirow{2}{*}{\parbox[t]{2.8cm}{isometric +\\orthographic (4x)}}
            & DeepCAD-vector & 81.51 & 75.14 & 23.40 & 11.37 \\
            & Drawing2CAD & \textbf{82.43} & \textbf{76.09} & \textbf{20.31} & \textbf{10.88} \\
        \bottomrule
    \end{tabular}
    }
\end{table}

\begin{table*}[t]
    \centering
    \small
    \caption{Ablation study on incremental versions of the model in three optional inputs. "dual dec." means dual-decoder architecture of our method. "guidance" means command-guidance operation used in parameters generation process. The concatenation-based learning embedding is added as the final component.}
    \label{tab:ablation}
    
    \begin{subtable}[t]{0.33\textwidth}
        \centering
        \captionsetup{font=footnotesize, justification=centering}
        \caption{isometric (1x)}
        \resizebox{\textwidth}{!}{
        \begin{tabular}{lcccc}
            \toprule
            \textbf{Method} & \textbf{ACC\textsubscript{cmd} $\uparrow$} & \textbf{ACC\textsubscript{param} $\uparrow$} & \textbf{IR $\downarrow$} & \textbf{CD $\downarrow$} \\
            \midrule
            DeepCAD-vector (baseline) & 80.78 & 73.73 & 23.29 & \textbf{11.52} \\
            dual dec. + baseline loss & \textbf{81.94} & 73.79 & 23.41 & 12.47 \\
            dual dec. + our loss & 81.86 & \textbf{74.47} & 23.88 & 11.87 \\
            dual dec. + our loss + guidance & 81.56 & 74.30 & 22.00 & 11.72 \\
            Drawing2CAD & 81.81 & 74.35 & \textbf{21.40} & 12.10 \\
            \bottomrule
        \end{tabular}}
    \end{subtable}
    \hfill
    \begin{subtable}[t]{0.33\textwidth}
        \centering
        \captionsetup{font=footnotesize, justification=centering}
        \caption{orthographic (3x)}
        \resizebox{\textwidth}{!}{
        \begin{tabular}{lcccc}
            \toprule
            \textbf{Method} & \textbf{ACC\textsubscript{cmd} $\uparrow$} & \textbf{ACC\textsubscript{param} $\uparrow$} & \textbf{IR $\downarrow$} & \textbf{CD $\downarrow$} \\
            \midrule
            DeepCAD-vector (baseline) & 81.39 & 74.76 & 22.97 & 12.15 \\
            dual dec. + baseline loss & 82.08 & 74.44 & 23.64 & 12.80 \\
            dual dec. + our loss & 82.10 & 75.17 & 22.59 & 12.51 \\
            dual dec. + our loss + guidance & 82.05 & 75.25 & 21.52 & 12.30 \\
            Drawing2CAD & \textbf{82.12} & \textbf{75.43} & \textbf{20.99} & \textbf{11.98} \\
            \bottomrule
        \end{tabular}}
    \end{subtable}
    \hfill
    \begin{subtable}[t]{0.33\textwidth}
        \centering
        \captionsetup{font=footnotesize, justification=centering}
        \caption{isometric + orthographic (4x)}
        \resizebox{\textwidth}{!}{
        \begin{tabular}{lcccc}
            \toprule
            \textbf{Method} & \textbf{ACC\textsubscript{cmd} $\uparrow$} & \textbf{ACC\textsubscript{param} $\uparrow$} & \textbf{IR $\downarrow$} & \textbf{CD $\downarrow$} \\
            \midrule
            DeepCAD-vector (baseline) & 81.51 & 75.14 & 23.40 & 11.37 \\
            dual dec. + baseline loss & 82.34 & 74.93 & 22.45 & 12.08 \\
            dual dec. + our loss & 82.28 & 75.56 & 22.43 & 11.23 \\
            dual dec. + our loss + guidance & 81.84 & 75.82 & 21.37 & 11.40 \\
            Drawing2CAD & \textbf{82.43} & \textbf{76.09} & \textbf{20.31} & \textbf{10.88} \\
            \bottomrule
        \end{tabular}}
    \end{subtable}
\end{table*}

\noindent
\textbf{Ablation Study.}
\label{ablation study}
We conducted ablation studies across three input configurations (single isometric view, three orthographic views, and all four views combined) to evaluate component contributions in Drawing2CAD, as shown in Table~\ref{tab:ablation}.
Starting with the DeepCAD-vector baseline, we incrementally added our proposed components: dual-decoder architecture, soft target distribution loss function, command-guided parameter generation, and concatenation-based embedding fusion.
With a single isometric view as input, our model achieves the lowest Invalid Ratio  while maintaining competitive performance in command accuracy, parameter accuracy, and Chamfer Distance. Validity is a particularly important metric in generative tasks, as it directly determines whether models can be used in practice. We argue that our approach strikes a careful balance among prediction accuracy, geometric fidelity, and structural validity, demonstrating robustness even under limited-view conditions.
For orthographic views (3x) and combined views (4x), Drawing2CAD significantly outperforms all reduced versions, indicating that each component in our framework is both essential and effective.
Notably, in the initial four reduced versions, we employed a linear addition strategy for embedding fusion, consistent with previous research. However, replacing this linear addition strategy with concatenation-based embedding fusion yielded significant performance improvement. These results confirm our hypothesis that concatenation-based embedding fusion better captures interdependencies among view, command, and parameter embeddings, creating more expressive and informative representations.

\section{Limitations and Discussions}
\label{sec:limitations}

\begin{figure}[t]
    \centering
    \includegraphics[width=\linewidth]{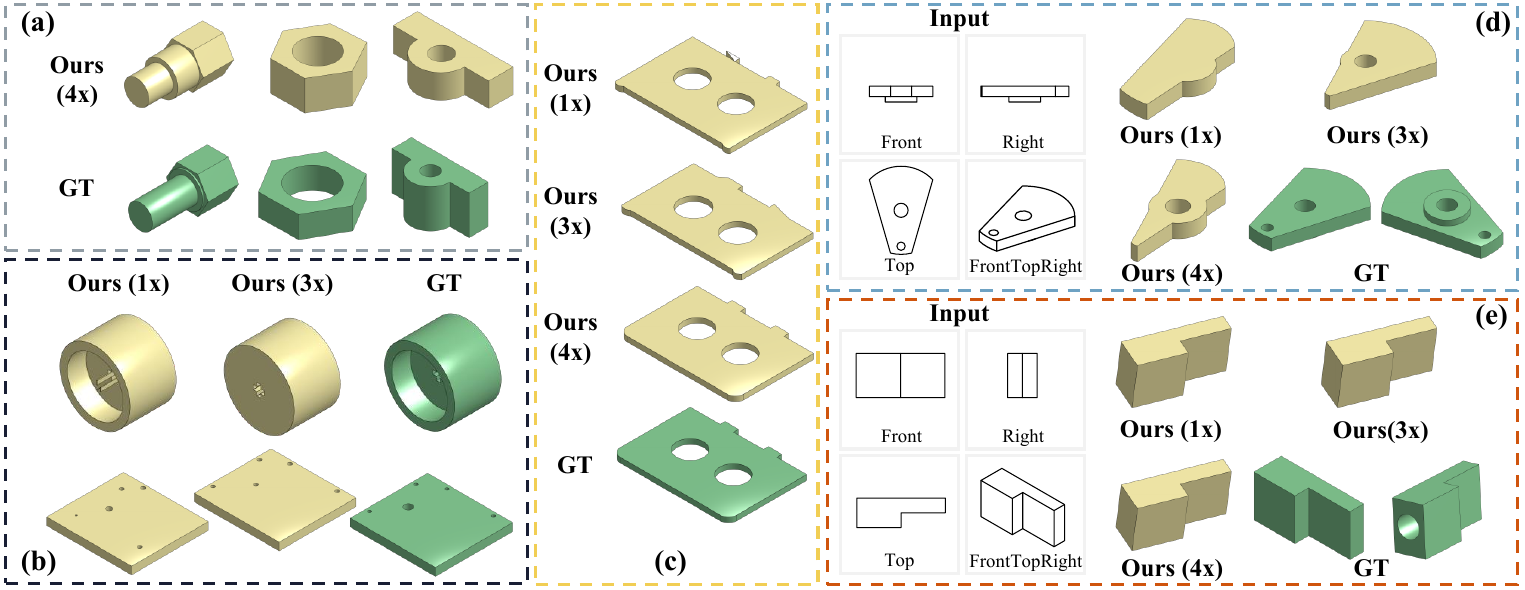}
    \caption{Five representative types of imperfect cases in our experiments which present limitations and insights of our method: Parameter Precision Issues (a), View-Specific Information Trade-offs (b), Multi-View Integration Challenges (c)(d), and View Information Dependency (e). "1x", "3x", "4x" means using a single isometric view, three orthographic views, or a combination of all four views as input respectively.}
    \label{fig:imperfect case}
\end{figure}

Despite its promising results in generating parametric CAD models from vector engineering drawings, our method still has room for improvement. Figure~\ref{fig:imperfect case} presents representative cases that highlight limitations and offer insights for future research. More details can be referred in the supplementary material.

\noindent
\textbf{Parameter Precision Issues.} As shown in Figure~\ref{fig:imperfect case} (a), our method may capture the overall shape yet show noticeable size deviations due to the tolerance design in the loss function. While this design stabilizes optimization, it can compromise parameter-level accuracy. This highlights the need for uncertainty-aware modeling to better balance precision and robustness.

\noindent
\textbf{View-Specific Information Trade-offs.} Different views introduce representational biases: orthographic views preserve symmetry or layout but lack depth, whereas isometric views enhance depth perception at the cost of planar alignment (Figure~\ref{fig:imperfect case} (b)). Future work could integrate geometric priors or neural rendering to compensate for such limitations and improve reconstruction consistency.

\noindent
\textbf{Multi-View Integration Challenges.} Multiple-view input enhances spatial completeness, but inconsistent cues such as the protrusion mismatch can lead to ambiguity (Figure~\ref{fig:imperfect case} (c), (d)). This suggests the need for robust multi-view fusion strategies to resolve visual conflicts and maintain structural coherence.

\noindent
\textbf{View Information Dependency.} Our approach fails to recover features invisible in all views, as demonstrated by the missing side hole in Figure~\ref{fig:imperfect case} (e). This limitation points to the potential of semi-supervised learning or implicit priors to infer occluded geometry and enrich the model’s ability to reconstruct hidden features.

\section{Conclusion}

\noindent
In this work, we propose Drawing2CAD, a novel approach for generating CAD operation sequences directly from vector engineering drawings, bridging the gap between 2D drawings and 3D CAD modeling. By redefining CAD model generation as a sequence-to-sequence learning task, our method leverages rich geometric information embedded in vector drawing sequences to produce CAD operation sequences that create functional CAD models.
Extensive experiments demonstrate that vector engineering drawings outperform raster inputs in command accuracy, parameter precision, and 3D reconstruction quality. Moreover, our method, with its tailored architecture and novel loss function, ensures effectiveness in CAD operation sequence generation. Additionally, we introduce CAD-VGDrawing, a large-scale dataset containing over 150,000 engineering drawings in both vector and raster formats, providing a valuable resource for future research in automated CAD design.
Our findings highlight the potential of integrating vector graphics and CAD operations in generative modeling, paving the way for more advanced, efficient, and intelligent CAD modeling frameworks.

\begin{acks}
National Natural Science Foundation of China (Nos. 62025207, 62072126), the Fundamental Research Funds for the Provincial Universities of Zhejiang (No. GK259909299001-006), and the Anhui Provincial Joint Construction Key Laboratory of Intelligent Education Equipment and Technology (No. IEET202401).
\end{acks}

\bibliographystyle{ACM-Reference-Format}
\bibliography{references}

\appendix
\twocolumn[
\begin{center}
    \Huge\bfseries Supplementary Material
     \vspace{1em}
\end{center}
]

\section{Imperfect Case Analysis}
As mentioned in Section~6, our method presents several opportunities for improvement. To better understand these opportunities, we present several representative types of imperfect cases and discuss them in the following sections.

\subsection{Parameter Precision Issues}
\begin{figure}[h]
  \centering
  \includegraphics[width=\linewidth]{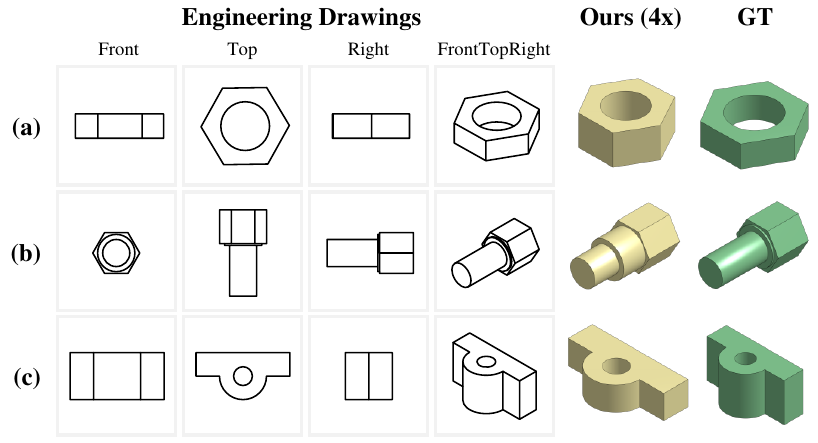}
  \caption{Imperfect cases about parameter precision issues. "4x" means using a combination of all four views as input.}
  \label{fig:ic-1}
\end{figure}

As shown in Figure~\ref{fig:ic-1}, the generated model generally matches the overall shape of the ground-truth, but size discrepancies remain due to the tolerance allowed in our loss function. In case (a), the model is thicker and the hole radius is smaller. Case (b) shows an overestimated disc thickness, while in case (c), the entire model is thinner than expected.

\subsection{View-Specific Information Trade-offs}
\begin{figure}[h]
  \centering
  \includegraphics[width=\linewidth]{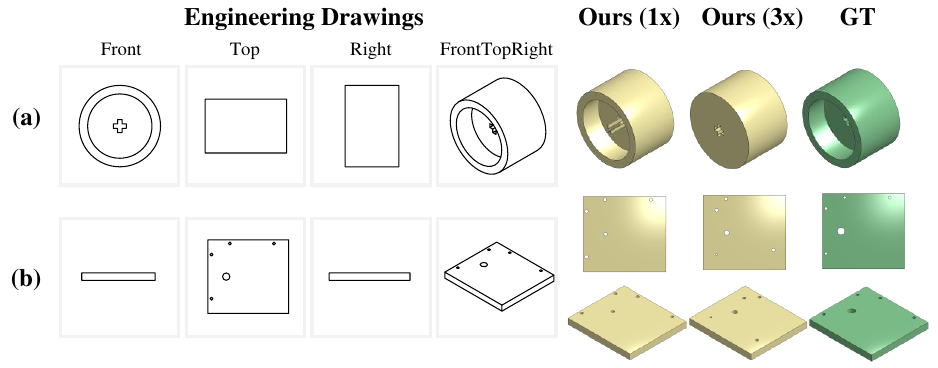}
  \caption{Imperfect cases about view-specific information trade-offs. "1x", "3x" means using a single isometric view or three orthographic views as input respectively.}
  \label{fig:ic-2}
\end{figure}

As illustrated in Figure~\ref{fig:ic-2}, three orthographic views help the model capture planar features like symmetry and layout, but often lead to inaccurate depth interpretation, as seen in case (a) where extrusion depth is misestimated. Conversely, the isometric view enhances depth perception but lacks precise planar alignment, resulting in disordered hole placement in case (b). These discrepancies highlight how different view types convey complementary geometric cues, influencing the model’s focus and generation behavior.

\subsection{Multi-View Integration Challenges}
\begin{figure}[h]
  \centering
  \includegraphics[width=\linewidth]{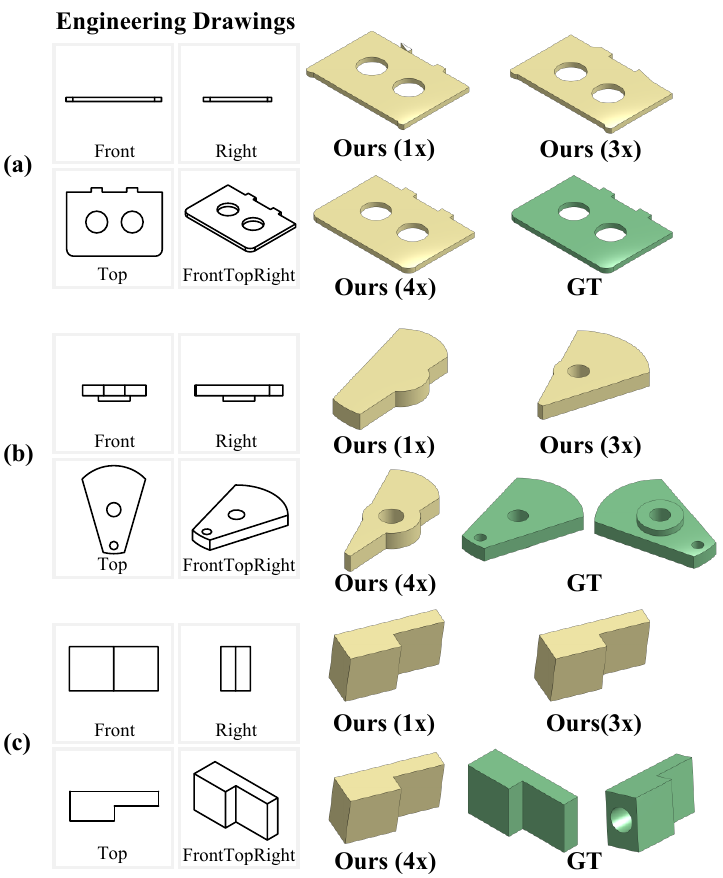}
  \caption{Imperfect cases about multi-view integration challenges. "1x", "3x", "4x" means using a single isometric view, three orthographic views, or a combination of all four views as input respectively.}
  \label{fig:ic-3}
\end{figure}

Combining all four views as input can provide complementary geometric cues, thereby enhancing the model’s overall understanding, as shown in Figure~\ref{fig:ic-3} (a). However, this strategy is not always reliable. When the information conveyed by different views is inconsistent, the model may receive conflicting signals, resulting in ambiguity and inaccurate reconstructions that deviate from the intended design. For instance, in Figure~\ref{fig:ic-3} (b), the orthographic views indicate a protrusion at the bottom of the object, while the isometric view omits this feature, leading to inconsistent guidance and an erroneous output.
In more extreme cases, such as the one shown in Figure~\ref{fig:ic-3} (c), none of the four views reveal a critical structural element—a hole on the left side of the model—due to occlusion or viewpoint limitations. Consequently, the generated model only captures the visible surfaces, failing to infer occluded or hidden geometry, which compromises both the completeness and the fidelity of the reconstruction.

\end{document}